\documentclass{ifacconf}

\usepackage{natbib}        
\usepackage[dvipsnames]{xcolor}
\usepackage{graphicx}
\usepackage{tabularx}
\usepackage{subcaption}
\usepackage{amsmath}
\usepackage{booktabs}
\usepackage{placeins}
\usepackage{graphicx}
\usepackage{caption}

\usepackage{graphics} 
\usepackage{soul}

\usepackage{tikz}
\usepackage{pgfplots}
\usetikzlibrary{positioning}
\usetikzlibrary{shapes.geometric}
\usetikzlibrary{shapes.misc}
\usetikzlibrary{shapes,arrows,patterns}
\usetikzlibrary{plotmarks}
\pgfplotsset{compat=newest}

\usepackage{paralist}
\usepackage{mathtools}

\usepackage{siunitx}
\usepackage{multirow}
\usepackage{epstopdf}
\graphicspath{ {Figures/} }
\usepackage{multicol}
\usepackage{epstopdf}

\usepgfplotslibrary{patchplots}
\usepackage{epsfig}
\usepackage{float}
\usepackage{amsmath} 
\usepackage{amssymb}  
\usepackage{mathtools}
\graphicspath{ {Figures/} }
\DeclareGraphicsExtensions{.pdf,.eps}

\DeclareMathOperator*{\argmin}{arg\,min}
\usepackage[acronym,nomain,nonumberlist]{glossaries}
\newacronym{ahe}{AHE}{Adaptive Histogram Equalization}
\newacronym{ar}{AR}{Augmented Reality}
\newacronym{clahe}{CLAHE}{Contrast Limited Adaptive Histogram Equalization}
\newacronym{cnn}{CNN}{Convolutional Neural Network}
\newacronym{dl}{DL}{Deep Learning}
\newacronym{dof}{DoF}{Degree of Freedom}
\newacronym{ekf}{EKF}{Extended Kalman Filter}
\newacronym{fps}{fps}{Frame Per Second}
\newacronym{fov}{FoV}{Field of View}
\newacronym{fbe}{FBE}{Forward Backward Envelope}
\newacronym{gnss}{GNSS}{Global Navigation Satellite System}
\newacronym{gps}{GPS}{Global Positioning System}
\newacronym{kf}{KF}{Extended Kalman Filter}
\newacronym{iid}{IID}{Independent Identically Distributed}
\newacronym{imu}{IMU}{Inertial Measurement Unit}
\newacronym{mae}{MAE}{mean absolute error}
\newacronym{mav}{MAV}{Micro Aerial Vehicle}
\newacronym{mhe}{MHE}{Moving Horizon Estimation}
\newacronym{mimo}{MIMO}{Multiple Input Multiple Output}
\newacronym{mlp}{MLP}{MultiLayer Perceptron}
\newacronym{mpc}{MPC}{Model Predictive Control}
\newacronym{msf}{MSF}{Multi Sensor Fusion}
\newacronym{nmhe}{NMHE}{Nonlinear Moving Horizon Estimation}
\newacronym{nmpc}{NMPC}{Nonlinear Model Predictive Control}
\newacronym{nn}{NN}{Nueral Network}
\newacronym{nwu}{NWU}{North-West-Up}
\newacronym{panoc}{PANOC}{Proximal Averaged Newton-type method for Optimal Control}
\newacronym{pdf}{PDF}{Probability Density Function}
\newacronym{pid}{PID}{Proportional Integral Derivative}
\newacronym{relu}{ReLU}{Rectified Linear Unit}
\newacronym{rmse}{RMSE}{Root Mean Square Error}
\newacronym{rl}{RL}{Reinforcement Learning}
\newacronym{ros}{ROS}{Robot Operating System}
\newacronym{sfm}{SfM}{Structure from Motion}
\newacronym{sqp}{SQP}{Sequential Quadratic Programming}
\newacronym{ugv}{UGV}{Unmanned Ground Vehicle}
\newacronym{ukf}{UKF}{Unscented Kalman Filter}
\newacronym{vi}{VI}{Visual Inertia}
\newacronym{vo}{VO}{Visual Odometry}
\newacronym{lidar}{LiDAR}{Light Detection And Ranging}
\newacronym{pcl}{PCL}{Point Cloud}
\newacronym{sub-t}{Sub-T}{Sub-Terranean}

\usepackage{bm}

\newlength\fwidth

\begin{document}
\begin{frontmatter}

\title{Autonomous Point Cloud Segmentation for Power Lines Inspection in Smart Grid} 


\author[First]{Alexander Kyuroson} 
\author[First]{Anton Koval} 
\author[First]{George Nikolakopoulos} 

\address[First]{Robotics \& AI Team, Department of Computer, Electrical and Space Engineering, Lule\r{a} University of Technology, Lule\r{a} SE-97187, Sweden.}

\begin{abstract}                
LiDAR is currently one of the most utilized sensors to effectively monitor the status of power lines and facilitate the inspection of remote power distribution networks and related infrastructures. To ensure the safe operation of the smart grid, various remote data acquisition strategies, such as Airborne Laser Scanning (ALS), Mobile Laser Scanning (MLS), and Terrestrial Laser Scanning (TSL) have been leveraged to allow continuous monitoring of regional power networks, which are typically surrounded by dense vegetation. In this article, an unsupervised Machine Learning (ML) framework is proposed, to detect, extract and analyze the characteristics of power lines of both high and low voltage, as well as the surrounding vegetation in a Power Line Corridor (PLC) solely from LiDAR data. Initially, the proposed approach eliminates the ground points from higher elevation points based on statistical analysis that applies density criteria and histogram thresholding. After denoising and transforming of the remaining candidate points by applying Principle Component Analysis (PCA) and Kd-tree, power line segmentation is achieved by utilizing a two-stage DBSCAN clustering to identify each power line individually. Finally, all high elevation points in the PLC are identified based on their distance to the newly segmented power lines. Conducted experiments illustrate that the proposed framework is an agnostic method that can efficiently detect the power lines and perform PLC-based hazard analysis.
\end{abstract}


\end{frontmatter}
\glsresetall
%
%
%
\section{Introduction}

Unmanned Aerial Vehicles (UAVs) have recently been widely deployed for monitoring and inspection of many infrastructures~\citep{Jordan2018}, such as bridges~\citep{Bono2022, Bolourian2020,Ameli2022}, wind turbines~\citep{Car2020}, as well as roads and construction sites~\citep{Shang2018} due to their low-cost, high-precision and efficiency~\citep{Liu2019,Wang2019}. Over the past few years, special interest has surged in the inspection of smart power grids and transmission networks due to the limitations of the traditional approaches and their safety and efficiency concerns~\citep{Teng2017}. High cost and human factors for managing vegetation in PLC can be additionally seen as the limiting aspects for currently utilized methods~\citep{Araar2015}. By increasing the autonomy during initial surveillance of power lines and surrounding vegetation encroachment into PLC, early detection of upcoming hazards and existing faults can be achieved with higher accuracy and frequency at a lower cost~\citep{Yang2020}. 

The long-term exposure to harsh environmental conditions, such as snow storms and lightning, as well as lack of access to regional smart power grids for continuous inspection, due to harsh terrain and dense forests, is one of the major challenges to ensure reliable and safe operation of transmission grids~\citep{Teng2017}. With the advancement in sensor-related technology, large-scale accessibility to Point Clouds (PCL)s and their utilization for remote data acquisitions, power line and vegetation patrolling by ALS and MLS can be further exploited to enhance the existing automated inspection frameworks for various man-made and natural constructions while monitoring their operational conditions based on the environmental factors~\citep{Wang2018}. The generated parametric models of such constructions, based on the point cloud map acquired from airborne LiDAR, can be used to reconstruct and assess these urban infrastructures and their topology to detect hazardous factors~\citep{Wang2018,Munir2019}.

\begin{figure*}
    \centering
    \includegraphics[width=0.96\textwidth]{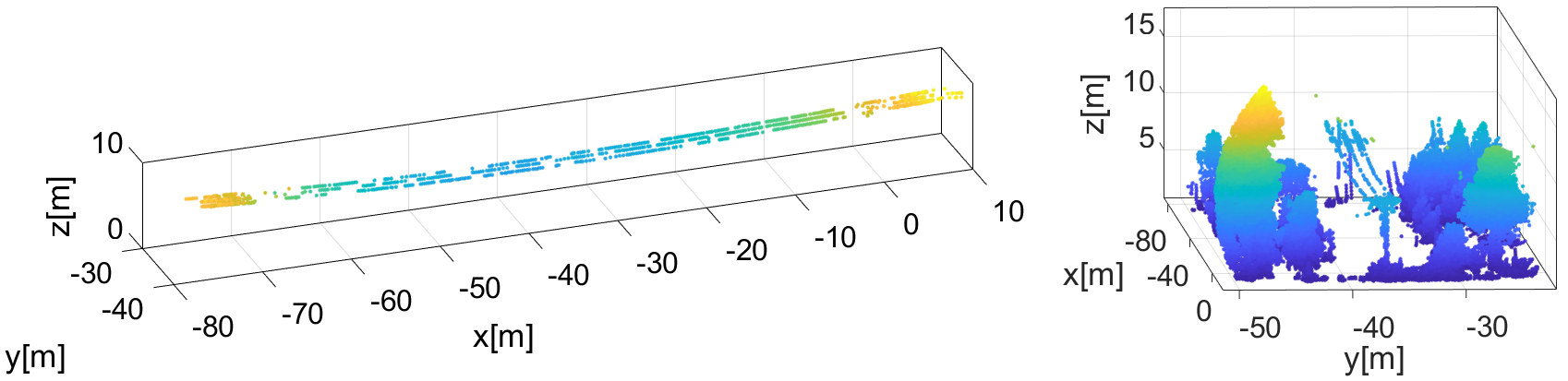}
    \caption{\textit{Left}: Extracted power lines from the LiDAR data. \textit{Right}: Extracted power line corridor.}
    \label{fig:Fig_1}
\end{figure*}

\subsection{Background \& Motivation}

Semantic segmentation of smart grids based on LiDAR data can be achieved by either utilizing supervised or unsupervised methods~\citep{Munir2019}. Pattern recognition techniques, such as Hough Transformation (HT)~\citep{Lin2016} and Random Sample Consensus (RANSAC) have been employed to search for parallel, linear structures~\citep{Teng2017} and 3D fit linear candidate points from the point cloud into segmented lines, while utilizing Markov Random Fields~\citep{Sohn2012}. A similar approach was proposed by~\citep{Bay2005} to exploit the topological relation between all the presented line segments. It must be noted that simultaneous multi-layered power line detection and segmentation based on HT in vertical arrangements in a harsh, noisy, and complex environment cannot be realized~\citep{Sohn2012}.

Furthermore, a method proposed by~\citep{McLaughlin2006} creates a segmentation of transmission lines by analyzing the covariance matrix of the LiDAR point cloud for each section in its ellipsoidal neighborhood. This enables the identification of linear and planar features, thereby allowing the recognition and classification of power lines, vegetation as well as other urban structures~\citep{Wang2018}. By employing statistical analysis based on the height, intensity, and distribution of LiDAR data,~\citep{Teng2017} proposed the detection and extraction of power line points, while performing rasterization. Similarly, a voxel-based piece-wise line detection method is proposed by~\citep{Jwa2009} to allow the reconstruction of power lines and their primitive model by unifying extracted features at various perceptual information layers during segmentation. Reconstruction of the individual power lines is achieved by a non-linear piece-wise fitting of segmented voxel space into the catenary curve equation~\citep{Wang2018, Jwa2009}. 

Other parametric approaches are generating a 3D model of power lines by utilizing Euclidean distance between seed candidate points in conjunction with quadratic polynomial fitting to analyze operational characteristics of transmission lines~\citep{Liang2011}. Dominant eigenvector fields resulting from the covariance matrix of a LiDAR point cloud are employed to assess the orientation of points in the occupancy map to extract candidate points, which are parallel to the axis-of-interest~\citep{Ritter2012}. Moreover, by assuming that neighboring transmission lines with similar characteristics are nearly parallel due to their connection to the same pylon, the RANSAC-based approach can be deployed to efficiently construct their primitive model~\citep{Guo2016, Cheng2014}. Iterative reconstruction and clustering of transmission lines based on rule-based M-estimator Sample Consensus (MSAC) in an urban environment with restrictive overhead inspection is proposed by~\citep{Cheng2014} to improve identification of power line points while addressing the sampling issues caused by nearby infrastructures. To obtain valid topographical relations based on MSAC, iterative calculation between each of the turning points is required to ensure accurate classification~\citep{Teng2017}.

The recent advancements in ML and specifically Deep Learning (DL) have provided an opportunity for their utilization to facilitate the semantic segmentation of smart grids~\citep{Munir2019, Qiao2020}. Point cloud segmentation for the detection of power lines based on K-means clustering of feature space is proposed by~\citep{Lin2016}. Additionally, other frameworks based on simple supervised learning algorithms such as Random Forest (RF)~\citep{Wang2021} and Joint-Boost classifiers~\citep{Guo2016, Guo2015}, have been proposed to classify power lines based on their geometric features, while employing graph-cut segmentation to optimize the results. Expectation Maximization (EM) has been used to implement a supervised parametric classification of airborne LiDAR data by merging feature sets from acquired LiDAR and images to classify urban structures~\citep{Lodha2007}. However, due to complex and continuously changing environments, the proposed methods have poor robustness and are limited in their usage~\citep{Yang2020}. This is mainly due to the reliance of such methods on effective feature design and selection during pre-processing of training data-set, which greatly impacts their real-life performance~\citep{Yang2020}. Although it is feasible to apply transfer learning to improve the feature expression, this approach cannot be applied to simple learning ML methods due to their requirement for small-scale data-set during their initial training and validation to achieve generalization~\citep{Yang2020}. 

More advanced supervised classifiers, such as SVM~\citep{Wang2017}, MLP~\citep{Zhao2016}, and DCNN~\citep{Yang2020}, have been utilized to overcome the detection challenges by relying on the big data and computation resources currently available. However, such classifiers required large training data sets and any unbalances in the distribution of samples provided in the training sets will result in bias and an increased rate of misclassification~\citep{Munir2019}. It has been shown that unsupervised learning methods, such as conditional Generative Adversarial Network (cGAN) proposed by~\citep{Gao2019} can achieve better performance and has better computation efficacy than Pix2pix~\citep{Isola2017} and SegNet~\citep{Badrinarayanan2017} models to inspect power insulators. Unsupervised DL methods have a great potential to provide automatic smart grid detection and analysis for an intelligent inspection~\citep{Yang2020}. However, due to insufficient and biased training samples, other reliable methods must be initially implemented to resolve overfitting issues~\citep{Qiao2020}. To accomplish such a task, fast and reliable unsupervised ML methods, such as DBSCAN, should be used to facilitate semantic segmentation for diverse environments, while incorporating an Explainable AI (XAI) framework to gain insight into the performance of the model during its operation for the collection of larger unbiased data-set~\citep{Abdollahi2021}. This approach is aimed at improving the existing models based on comprehension of the algorithm for power line and PLC inspection, which can lead to further optimization of supervised DL methods~\citep{Yang2020, Qiao2020}, thereby overcoming some shortcomings and limitations of existing deep architectures such as PointNet++~\citep{Nurunnabi2021} and GCN~\citep{Li2022}.

\subsection{Contributions}

The contributions of this article can be summarized as follows:

\begin{enumerate}
    \item A novel two-stage unsupervised hierarchical clustering that allows the identification and segmentation of individual power lines by proposing simultaneous utilization of DBSCAN~\citep{Easter1996} and Kd-tree to handle the outliers and dense point clouds.
    \item A new modular agnostic pipeline that can achieve state-of-the-art performance in the detection and segmentation of individual power lines solely from unorganized and unlabeled LiDAR data. 
    \item The proposed approach has an intrinsic on-the-fly outlier detection and removal in harsh environments during power line inspection based on the proposed combination of statistical methods, PCA, and histogram threshold, as well as density-based clustering.
    \item The unsupervised modular approach fits easily in the XAI framework~\citep{Abdollahi2021} and does not require any training with labeled LiDAR data prior to its deployment in contrast to PointNet++~\citep{Nurunnabi2021} and GCN~\citep{Li2022}, while its performance can be easily assessed and improved given any newly introduced criteria for semantic segmentation of PLC. 
\end{enumerate}

\subsection{Outline}

The remainder of this article is structured as follows. Section~\ref{sec:pipeline} describes the newly proposed approach while establishing each module and its role in the segmentation. Section~\ref{sec:result} presents the experimental setup with achieved results. Finally, Section~\ref{sec:conclusion} provides the concluding remarks and discusses the achieved results.

\section{Segmentation Pipeline}
\label{sec:pipeline}

\begin{figure*}[!t]
\begin{center}
    \includegraphics[width=1\linewidth, trim={2.5cm 17.7cm 1.0cm 5.6cm}, clip, page=1]{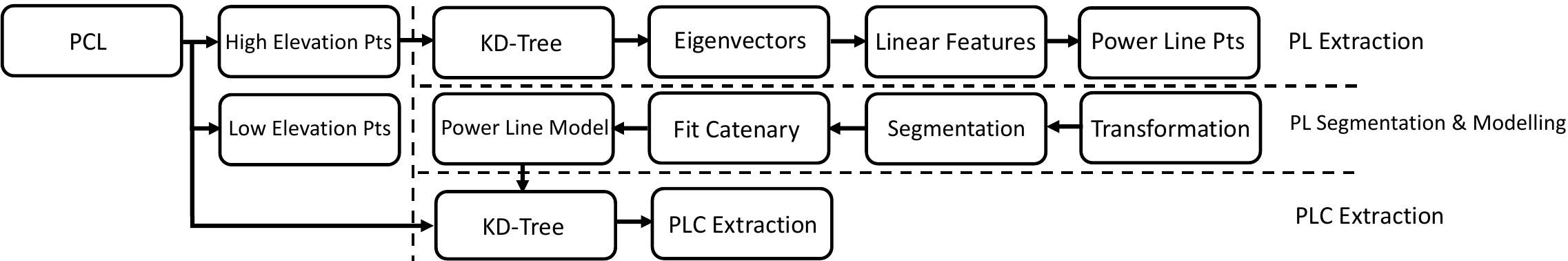}
    \caption{The schematics of the proposed pipeline for extraction of power lines and Power Line Corridor (PLC).
    \label{f:Fig_2}}
\end{center}
\end{figure*}

In this Section, the proposed two-stage unsupervised hierarchical clustering method based on DBSCAN and Kd-tree for extraction of power lines and the PLC is presented. The overall proposed pipeline with its subsequent processes is shown in Fig.~\ref{f:Fig_2}. The proposed framework solely requires unorganized unlabeled point clouds and their $xyz$-coordinate values, which are obtained during overhead power line inspection by a UAV. Moreover, any additional data is discarded prior to continuation to minimize the impact on the computational resources. Overall, the pipeline can be divided into five major segments, high elevation filtration, power line extraction as well as segmentation, power line parametric modeling and characterization, and PLC extraction to investigate hazardous conditions. The details of each process are presented in the sequel.

\subsection{High Elevation Extraction}

To improve the efficiency of the proposed pipeline for the detection and extraction of power lines, effective outlier removal and ground point filtration are crucial~\citep{Huang2021}. It must be noted that more than 90\% of the existing points in any 3D point cloud collected by UAVs constitute the surrounding terrain and vegetation~\citep{Li2022}. Based on the height distribution of the LiDAR data, the height histogram was employed to extract high elevation points. Similar approaches have been adopted successfully by~\citep{Huang2021,Shi2018} to reduce the ground points while maintaining points that are corresponding to the power lines and their surrounding vegetation. 
In this paper, the method proposed by~\citep{Shi2018} is used to extract high elevation points from the initial data to reduce computational complexity. Moreover, the number of equally spaced bins, as well as the height threshold, $\beta$, for the presented data is determined based on hyper-parametric grid optimization, which is presented in the upcoming subsection.

\subsection{Power Line Extraction \& Regularization}


Extracted high-elevation points contain various types of objects, such as power lines, surrounding vegetation as well as electrical poles. To distinguish power lines in the airborne 3D LiDAR data, PCA is utilized to evaluate eigenvalue-related features~\citep{Wang2021}. The Linear-likeness (LN) based on the eigenvalues can be calculated by utilizing the covariance matrix of 3D coordinates of each point through eigenvalue decomposition:

\setlength{\abovedisplayskip}{-12pt}
\setlength{\belowdisplayskip}{0pt}
\setlength{\abovedisplayshortskip}{0pt}
\setlength{\belowdisplayshortskip}{0pt}
\begin{align} \label{e:Eq_1}
    (\Sigma-\lambda I) \overrightarrow{v} = 0,
\end{align}

where $\lambda$ and $\overrightarrow{v}$ are the eigenvalues and eigenvectors respectively, $\Sigma$ is a covariance matrix defined as:

\setlength{\abovedisplayskip}{-12pt}
\setlength{\belowdisplayskip}{0pt}
\setlength{\abovedisplayshortskip}{0pt}
\setlength{\belowdisplayshortskip}{0pt}
\begin{align} \label{e:Eq_2}
    \Sigma = \frac{1}{N} (\tilde{X^T}\tilde{X}),
\end{align}

where $\tilde{X}$ and $N$ represent the mean-centered data and instances of points in the LiDAR data, respectively~\citep{Burt2019}. Furthermore, it is vital to use Kd-tree to structure and spatially organize the PCL prior to the calculation of the covariance matrix such that $\Sigma$ only contains points belonging to the transmission line for an accurate extraction~\citep{Shi2018}. The equation for linear features based on calculated eigenvalues is as follows:

\setlength{\abovedisplayskip}{-10pt}
\setlength{\belowdisplayskip}{0pt}
\setlength{\abovedisplayshortskip}{0pt}
\setlength{\belowdisplayshortskip}{0pt}
\begin{align} \label{e:Eq_3}
    LN = \frac{\lambda_1 - \lambda_2}{\lambda_1},
\end{align}

where $\lambda_1$ and $\lambda_2$ are the calculated eigenvalues, which must satisfy the condition, $\lambda_1\gg\lambda_2\approx\lambda_3$. Therefore, the $LN$ for linear components must converge to $1$, while its corresponding eigenvector, $\overrightarrow{v_1}$, is approximately parallel to $Z$-axis~\citep{Guan2016}. Given that these two conditions are satisfied for any given point in the point cloud while having more than $4$ nearest neighbors, they are classified as power line candidates. Other eigenvalue features, such as planar-likeness (PL) and spherical-likeness (SP) can be calculated to differentiate vegetation and buildings, respectively~\citep{Wang2021}.

Given the inherent capability of DBSCAN to identify and remove un-clustered observations as noise~\citep{Bryant2018}, it is feasible to directly feed the extracted power line candidate points into the two-stage clustering algorithm. However, to ensure better quality during segmentation and decrease computational time during hyper-parametric optimization, a simple outlier removal is employed prior to the segmentation of transmission lines. To remove outlier points from the remaining candidates, statistical analysis is used to calculate the mean distance~\citep{Shi2018}. 
Furthermore, by utilizing the previously calculated eigenvectors, $\overrightarrow{v_1}$ and $\overrightarrow{v_2}$, the angle between the vertical plane of transmission line and $ZY$-plane can be measured as $\theta = \arctan(\overrightarrow{v_1}/\overrightarrow{v_2})$, where $\theta$ is the rotation angle around $Z$-axis~\citep{Huang2021}. To validate the resulted $\theta$, the angle between $ZY$-plane and the vertical plane of the power line is evaluated by $\theta = \arccos (\hat{n}_1.\hat{n}_2/|\hat{n}_1||\hat{n}_2|)$, where $\hat{n}_1$ and $\hat{n}_2$ are unit normals to $ZY$-plane and the vertical plane, respectively. The homogeneous rotation matrix $R_z$~\citep{corke2011robotics} is used to regularize the point cloud, thereby simplifying clustering and parametric modeling of the segmented power lines; see Fig.~\ref{fig:Fig_1}. 



\subsection{Power Line Segmentation}

To achieve an agnostic semantic segmentation of individual power lines with various configurations in complex environments, two-layered sequential DBSCAN is implemented to analyze the direction of the power line span along both the orthogonal and parallel planes. 
Moreover, DBSCAN is selected as the preferred clustering method due to its application for both convex and non-convex sample sets~\citep{Chi2019}.

To implement DBSCAN, $eps$ and $min_{pts}$ are selected such that the number of resulted clusters is minimized, while their density is maximized. In other words, $eps$, which represents the radius of clusters from each centroid, $C_i$, for any given set of observations, is minimized based on the single linkage principle to maximize the similarity between the clustered points around $C_i$, while the cluster size expressed as $min_{pts}$, is maximized based on the complete linkage principle to minimize the similarities between the resulted clusters. 

The proposed approach ensures the creation of denser clusters and robust identification of observation points as noise~\citep{Bryant2018}. Furthermore, it is feasible to adopt a more simplified DBSCAN variation, RNN-DBSCAN, proposed by~\citep{Bryant2018} to automate the parameter selection, while utilizing divisive clustering to improve the latency and computational complexity. On contrary, the goal of our method is mainly to address the two main drawbacks of single-layered solo DBSCAN. These include parameter selection and optimization, as well as difficulty distinguishing clusters with high-density variation in the 3D point cloud. Moreover, in our study, hierarchical clustering principles are utilized during the hyper-parametric optimization of DBSCAN to assess the approach and its efficacy.

\subsection{Power Line Parametric Modelling}

To assess the operational status and safety of the power lines after their segmentation, their geometrical characteristics based on the catenary curve can be modeled as follows:

\setlength{\abovedisplayskip}{-15pt}
\setlength{\belowdisplayskip}{0pt}
\setlength{\abovedisplayshortskip}{0pt}
\setlength{\belowdisplayshortskip}{0pt}
\begin{align} \label{e:Eq_6}
    y(x)= a + c\cosh (\frac{x-b}{c}),
\end{align}

where $a$, $b$ and $c$ represent vertical displacement from the $X$-axis $[m]$, horizontal displacement from the $Y$-axis $[m]$ and vertical position of vertex for the catenary, respectively~\citep{Yermo2019}. Moreover, the parameter $c$ can be formulated by $c= \frac{T_0}{w}$, where $T_0$ $[N]$ presents the tension at the sag position and $w$ $[N/m]$ is the linear weight density of the power line. By employing the least-squares, parameters $a$, $b$ and $c$ can be solved, using the following equation:

\begin{align} \label{e:Eq_7}
    &\argmin_{a,b,c} \frac{1}{n}\sum_{i=1}^{n} \big [ y^{'}-c\cosh(\frac{x^{'}-b}{c}+a) \big ],
\end{align}

where $x^{'}$ and $y^{'}$ are the $y$- and $z$-coordinates of any given power line points after regularization~\citep{Huang2021}. It has been shown~\citep{Yermo2019} that in many cases, the catenary curve equation can be simplified and expressed by the quadratic formula, $y(x) = a_2x^2+a_1x+a_0$, where $a_2$, $a_1$ and $a_0$ are coefficients of the fitted power line in $ZY$-plane~\citep{Liang2011}. This approximation allows faster evaluation of the condition of the transmission line and employment of MSAC to eliminate any remaining outliers that can affect the parametric modelling~\citep{Cheng2014}. Since the increase in the sag of transmission lines during winter time due to the build-up of snow and icicles can overload the power line, which can cause instabilities and irreparable damage to the transmission line, residual assessment of the polynomial fitting can easily detect hazardous conditions.

\subsection{PLC Extraction}

After segmentation and identification of each power line, PLC is extracted by using Kd-tree to achieve a higher neighboring point locality with the sole aim of improving overall performance and minimizing the impact on the memory~\citep{Yermo2019}. The search radius, $r$, is selected based on the following conditions:

\setlength{\abovedisplayskip}{-5pt}
\begin{equation} \label{e:Eq_9}
  r =
    \begin{cases}
      max \{\Delta z, d_i \} & \text{if the environment is complex}\\
      max \{a_0\}                    & \text{otherwise},
    \end{cases}       
\end{equation}

where $\Delta z$ is a distance between any point $q$ and 2D catenary curve in the $ZY$-plane and $d_i$ represent the maximum distance of any point $P=(x, y, z)$, from the conductor plane, $V$. Provided that the $z$-coordinate value for the point $P$ is larger than its $d_i$, which is calculated by Pythagorean theorem~\citep{Huang2021}, the maximum value is selected for $r$. If the surrounding environment does not contain any structures or dense vegetation that are higher than the height of the pylons, $a_0$, which expresses the height of sag in $ZY$-plane, is chosen as $r$, to assess any vegetation under transmission lines with a substantial risk of interference. To extract the PLC and evaluate hazardous objects closest to transmission lines, a classification approach is utilized.

\subsection{Hyper-parametric Optimization}

To optimize the proposed framework for any given LiDAR data, an initial grid-based hyper-parametric optimization is required to ensure that the semantic segmentation is performed accurately. As previously discussed, the parameters for high elevation filtration, as well as Kd-tree and DBSCAN are terrain specific, and the configuration and length of the surveyed transmission lines will affect their performance. Therefore, based on the assumption that there exists at least one power line in the provided point cloud, a grid-based optimization is performed for all the parameters shown in Table~\ref{t:Tab_1}. To eliminate any outliers during the segmentation, the sets of criteria are tuned as a scale-able constraint to avoid any over-segmentation, as shown in Table~\ref{t:Tab_1}. 

\begin{table}[!ht]
\setlength{\abovecaptionskip}{1pt}
\setlength{\belowcaptionskip}{-12pt}
\caption{Required parameters to be optimized for the proposed algorithm.}\label{t:Tab_1}
\centering
\begin{tabular}{p{1.4cm}p{1cm}p{5.1cm}l|c|l}\toprule
        \textbf{Parameters}    &{\textbf{Initial Values}} &{\textbf{Hyper-parametric Optimisation Conditions}}  \\ \hline \hline
        $nbin$          &{100}   &{$\min_{nbin} = \{nbin | nbin < 500\}$}\\ \hline
        $\beta$         &{4}     &{$\min_\beta = \{\beta | \beta > 3\}$}\\ \hline
        $r$             &{0.6}   &{$\max_r = \{r | r > a_0 \}$}\\ \hline
        $LN_{thres}$    &{0.82}  &{$\max_{LN_{thres}} = \{LN_{thres} | LN_{thres} \in [0.5, 1] \}$} \\ \hline
        $\alpha_{thres}$&{5}     &{$\min_{\alpha_{thres}} = \{\alpha_{thres} | \alpha_{thres} \in [0, 20] \}$}\\ \hline
        $Filt_{thres}$  &{0.8}   &{$\min_{Filt_{thres}} = \{Filt_{thres} | Filt_{thres} > (\mu +4\sigma)\}$,~\citep{Shi2018}}\\ \hline
        $eps$           &{0.2}   &{$\min_{eps} = \{eps | eps > 0.1\}$}\\ \hline
        $min_{pts}$    &{50}     &{$\max_{min_{pts}} = \{min_{pts} | min_{pts} > 30\}$}\\ \hline \bottomrule
     \end{tabular}
\end{table}

\section{Results} 
\label{sec:result}

In this Section, the experimental platform and the acquired results based on the surveyed power line are presented. The current LiDAR data was collected by utilizing an autonomous flight over the power lines in Lule\r{a}, Sweden.

\subsection{Experimental Setup} 
\label{sec:setup}

\begin{figure} [t!] \centering
\includegraphics[width=1\linewidth]{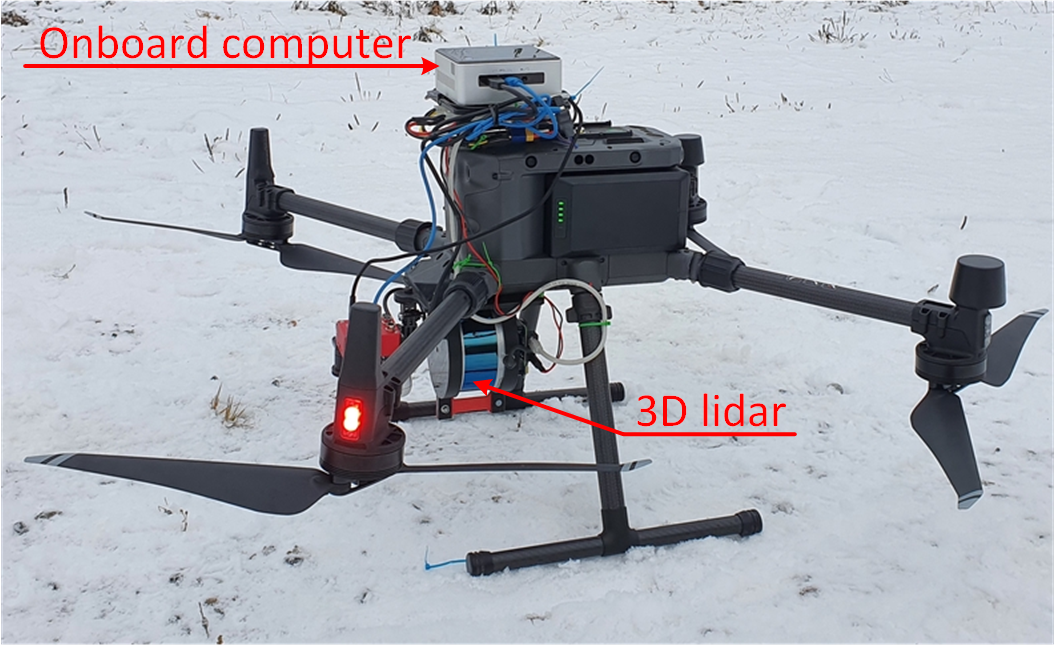}
\caption{Aerial system for power line inspection.}
\label{fig:Fig_7}
\end{figure}

\begin{figure} [ht!] \centering
\includegraphics[width=1\linewidth]{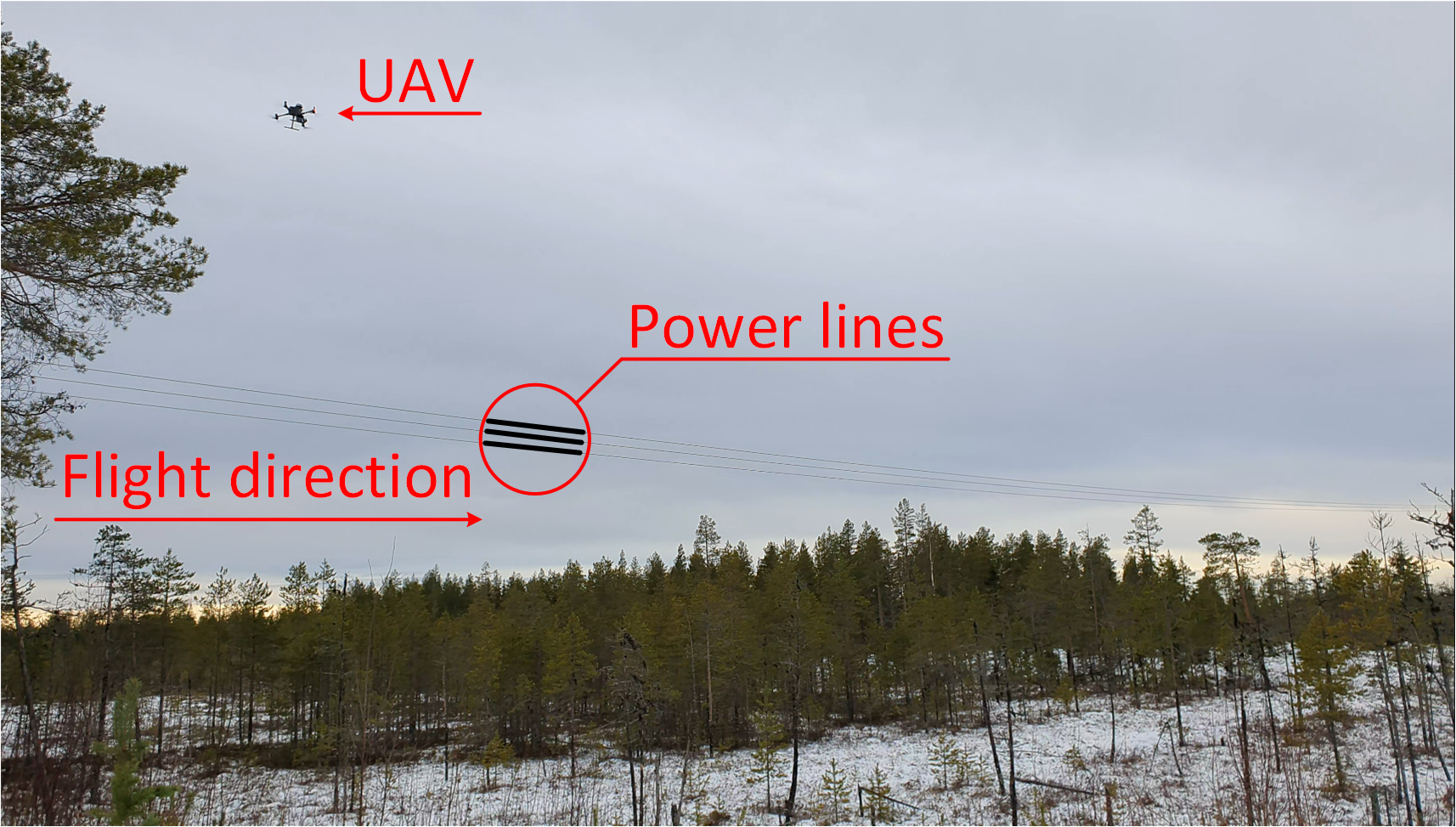}
\caption{Autonomous inspection of power line with UAV.}
\label{fig:Fig_8}
\end{figure}

As shown in Fig.~\ref{fig:Fig_8}, a UAV equipped with an onboard computer and 3D LiDAR is deployed to investigate the power lines and their surrounding vegetation. The onboard computer was equipped with Intel NUC with an i7 CPU, 16 GB of RAM, and SSD storage. The equipped LiDAR is Velodyne Puck Lite that can provide a 100 meters range with 360-degree $\times$ 30-degree horizontal and vertical fields of view and is capable of publishing data at 10 Hz, as shown in Fig.~\ref{fig:Fig_7}.

\subsection{Experimental Evaluation} 
\label{sec:evaluation}

\begin{figure} [!b] \centering
\includegraphics[width=1\linewidth]{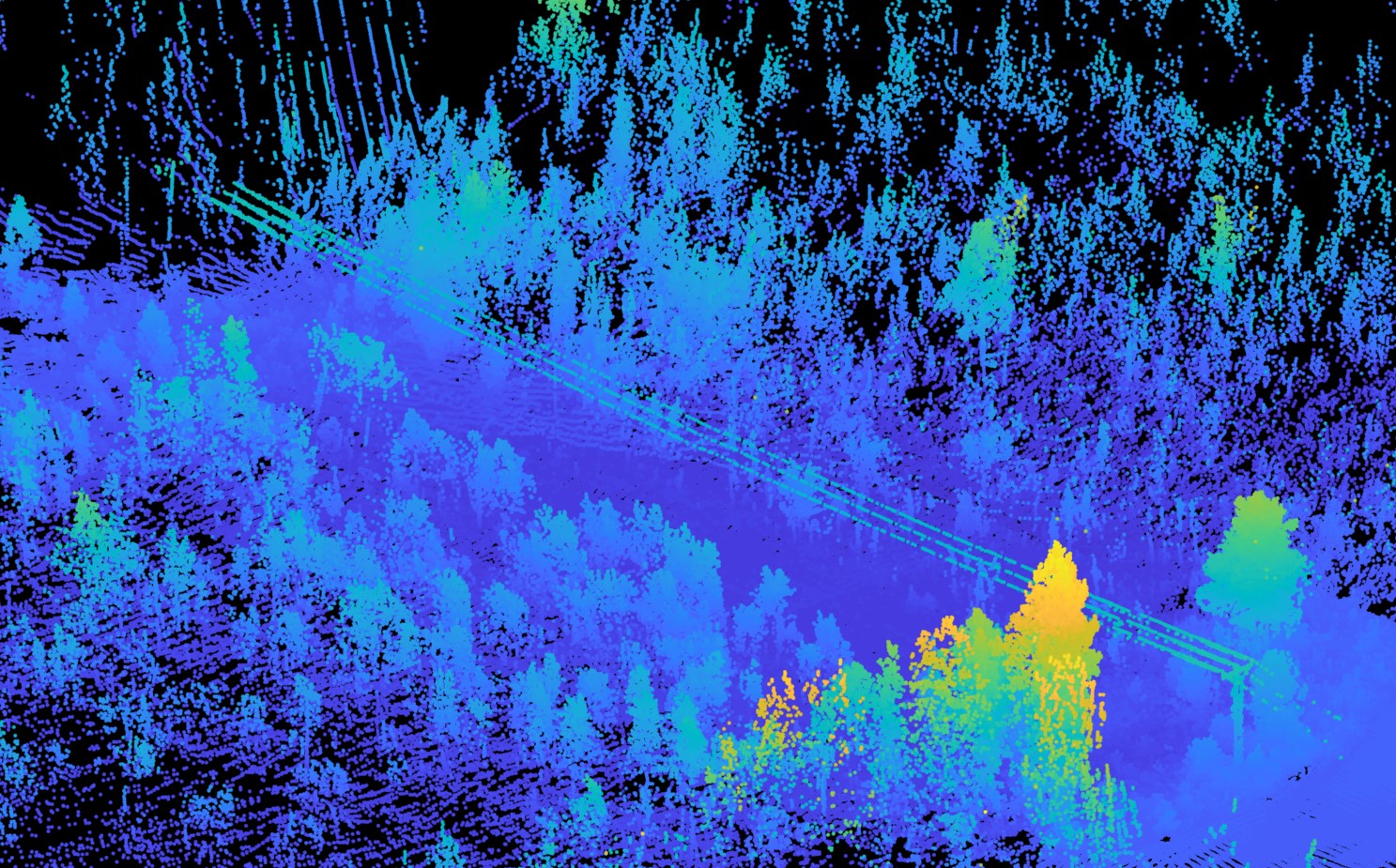}
\caption{3D Point Cloud map of the inspected area.}
\label{fig:Fig_9}
\end{figure}

\begin{figure*}
     \centering
     \begin{subfigure}[t!]{0.75\linewidth}
         \centering
         \includegraphics[width=1\linewidth]{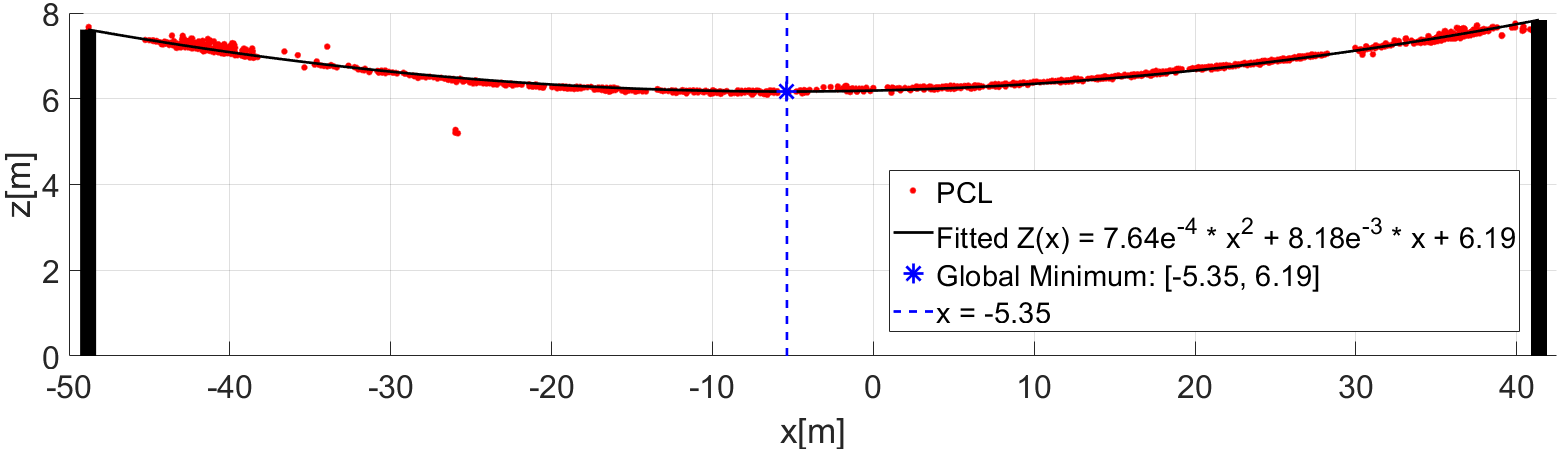}
         \caption{Power lines modeling}
         \label{fig:Fig_4a}
     \end{subfigure}
     \begin{subfigure}[t!]{0.75\linewidth}
         \centering
         \includegraphics[width=1\linewidth]{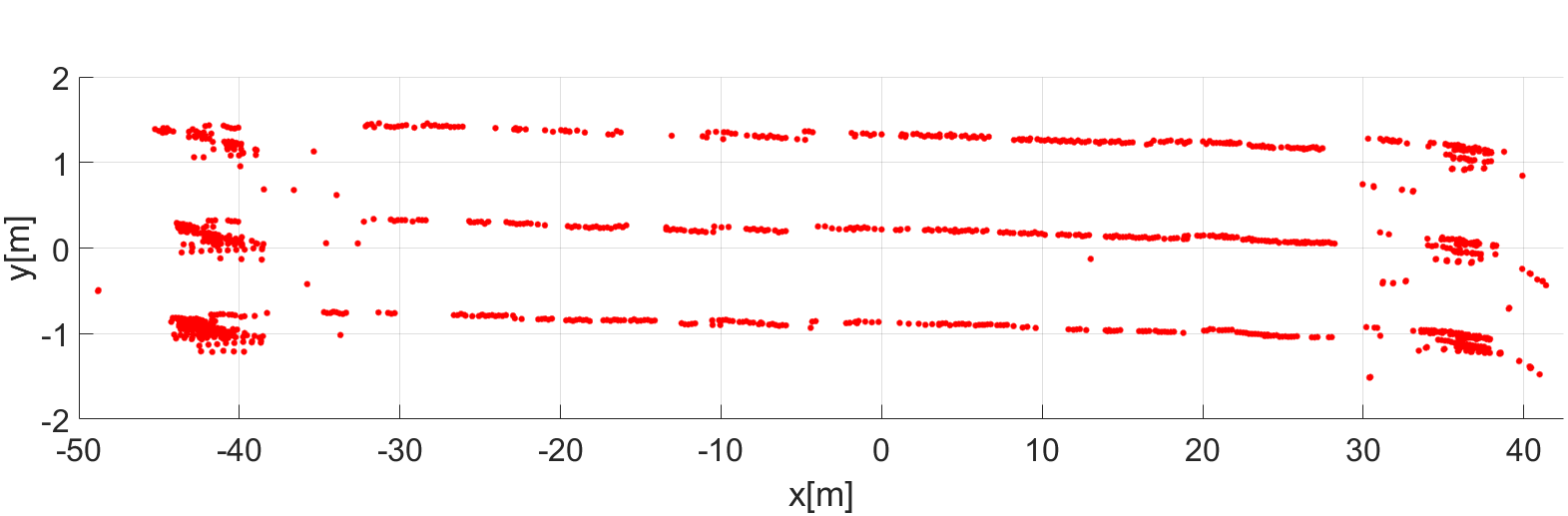}
         \caption{Power lines projected on $XY$-plane}
         \label{fig:Fig_4b}
     \end{subfigure}
     \begin{subfigure}[t!]{0.75\linewidth}
         \centering
         \includegraphics[width=1\linewidth]{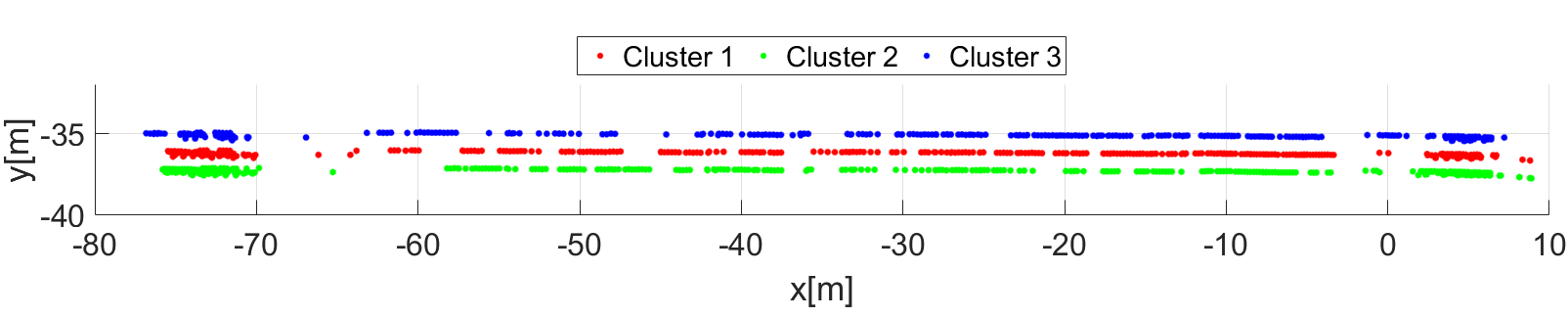}
         \caption{Segmented power lines projected on $XY$-plane.}
         \label{fig:Fig_4c}
     \end{subfigure}
    \setlength{\abovecaptionskip}{-10pt}
    \setlength{\belowcaptionskip}{-25pt}
     \caption{Power line extraction, segmentation, and modeling of each conductor for operational status assessment.}
     
     \label{fig:Fig_4}
\end{figure*}

By using the unprocessed LiDAR data from the UAV shown in Fig.~\ref{fig:Fig_9}, the power lines can be extracted as depicted in Fig.~\ref{fig:Fig_1} by the proposed framework. The initial height histogram analysis allowed the filtration of approximately $50\%$ of points associated with ground, as well as some outliers with low elevation. Afterward, linear feature extraction based on Kd-tree removed the remaining vegetation and poles. As shown in Fig.~\ref{fig:Fig_1}, the extracted transmission lines are regularized to facilitate their characterization and segmentation. The parametric modeling of extracted power lines is performed by using polynomial fitting as described previously to not only investigate the sag of power lines but also detect the position of the poles; see Fig.~\ref{fig:Fig_4a}. Fig.~\ref{fig:Fig_4b} illustrates the regularized extracted power lines, while projected on the $XY$-plane prior to segmentation. This enables inspection of possible interference and contact between the neighboring conductors on the same span, which can be used to characterize damages to the insulator due to mechanical overload that is  resulted from high applied shear forces, which is mainly caused by the built-up of icicles on the power lines during winter. Furthermore, by comparing Fig.~\ref{fig:Fig_4b} and Fig.~\ref{fig:Fig_4c}, the viability of the proposed algorithm based on two-stage DBSCAN is demonstrated and its capability in outlier removal is presented. Due to the utilization of a global 3D map, generated from the LiDAR scans and UAV odometry, from power lines during semantic segmentation without any regularization, the $x$ and $y$ values differ between Fig.~\ref{fig:Fig_4b} and Fig.~\ref{fig:Fig_4c}. 

Although the presented configuration of the power lines is simple in nature, the proposed method can assess multi-layered complex power line configurations, where several power lines are located both vertically and horizontally in harsh environments. As illustrated in Fig.~\ref{fig:Fig_1}, PLC extraction is achieved by using Kd-tree, and the optimal search radius, $r$, is selected as stated in Table~\ref{t:Tab_1} so that any interference from the surrounding vegetation can be detected. As depicted in the previous section, modular implementation of our pipeline enables the assessment of individual steps, thereby adhering to the XAI framework, where the functionality of each component can be easily explained and investigated for troubleshooting, as well as possible future improvements for the on-the-fly implementation.

As illustrated in Table~\ref{t:Tab_2}, the runtime of our method post its initial hyperparameter optimization is directly compared with the pre-trained built-in PointNet++ implementation of LiDAR semantic segmentation in Matlab. Although this comparison does not account for the initial training and validation phase of the PointNet++ and the optimization of the proposed pipeline, it demonstrates its potential for utilization in the autonomous inspection of power lines.

\begin{table}[H]
\caption{Runtime comparison between PointNet++ and our method.}\label{t:Tab_2}
\centering
\scalebox{0.82}{
\begin{tabular}{lcccc}\toprule
        \textbf{Methods}        &\multicolumn{2}{c}{\textbf{Runtime} [$ms/m^2$]} &\multicolumn{2}{c}{\textbf{PCL Density}}  \\ \hline 
        $ $              &\multicolumn{1}{|c|}{High Voltage} &\multicolumn{1}{c|}{Low Voltage} &\multicolumn{1}{|c|}{High Voltage} &\multicolumn{1}{c}{Low Voltage} \\ \hline \hline
        $PointNet++$     &\multicolumn{1}{|c|}{0.93} &\multicolumn{1}{c|}{0.87} &\multicolumn{1}{|c|}{10000429} &\multicolumn{1}{c}{928146} \\ \hline
        $Our Method$    &\multicolumn{1}{|c|}{0.25} &\multicolumn{1}{c|}{0.21} &\multicolumn{1}{|c|}{10000429} &\multicolumn{1}{c}{928146} \\ \hline \bottomrule
     \end{tabular}}
     
\end{table}

\subsection{Limitations}
\label{sec:limitations}
Compared to existing methods, the proposed pipeline has three major limitations: (a) It does rely on an initial hyperparameter optimization, which takes on average 15 minutes for any given new configuration of smart power grids, due to a lack of prior knowledge of PCL distribution. (b) It is sensitive to extrinsic factors such as occlusion and distortion of the dataset, which can be partially resolved by utilizing the model fitting to approximate the power line characteristics, thereby overcoming this limitation. (c) It requires multi-stage segmentation to leverage geometrical feature assessment to remove the noise and propagated errors in the PCL.

\section{Conclusions}
\label{sec:conclusion}
In this article, an agnostic unsupervised hierarchical clustering method to automatically detect, extract and characterize power lines is presented. An effective and fast algorithm based on histogram analysis and geometrical feature selection and filtration is proposed to select power line candidates. The proposed candidate detection is based on Kd-tree and demonstrated satisfactory performance by distinguishing power lines from surrounding vegetation and ground. After extraction of power lines, semantic segmentation is performed by the proposed two-stage DBSCAN to split and identify different power lines, thereby facilitating parametric modeling and assessment of an individual power line. To optimize the proposed framework, grid-based optimization is performed. However, evolutionary algorithm-based optimization can be employed to improve the overall performance during the tuning phase. Moreover, other variations of DBSCAN, such as HDBSCAN~\citep{McInnes2017} and RNN-DBSCAN~\citep{Bryant2018} for segmentation of the power lines can be further evaluated and their performance, when compared to the current framework, investigated to achieve better clustering, while the computational impact is minimized. Finally, the approach can be used to monitor and isolate the vegetation located in PLC for possible hazardous conditions.

%
\bibliography{main}

\begin{thebibliography}{4}
\providecommand{\natexlab}[1]{#1}
\providecommand{\url}[1]{\texttt{#1}}
\providecommand{\urlprefix}{URL }
\expandafter\ifx\csname urlstyle\endcsname\relax
  \providecommand{\doi}[1]{doi:\discretionary{}{}{}#1}\else
  \providecommand{\doi}{doi:\discretionary{}{}{}\begingroup
  \urlstyle{rm}\Url}\fi

\bibitem[{Able(1956)}]{Abl:56}
Able, B. (1956).
\newblock Nucleic acid content of microscope.
\newblock \emph{Nature}, 135, 7--9.

\bibitem[{Able et~al.(1954)Able, Tagg, and Rush}]{AbTaRu:54}
Able, B., Tagg, R., and Rush, M. (1954).
\newblock Enzyme-catalyzed cellular transanimations.
\newblock In A.~Round (ed.), \emph{Advances in Enzymology}, volume~2, 125--247.
  Academic Press, New York, 3rd edition.

\bibitem[{Keohane(1958)}]{Keo:58}
Keohane, R. (1958).
\newblock \emph{Power and Interdependence: World Politics in Transitions}.
\newblock Little, Brown \& Co., Boston.

\bibitem[{Powers(1985)}]{Pow:85}
Powers, T. (1985).
\newblock Is there a way out?
\newblock \emph{Harpers}, 35--47.

\end{thebibliography}


\begin{thebibliography}{44}
\providecommand{\natexlab}[1]{#1}
\providecommand{\url}[1]{\texttt{#1}}
\providecommand{\urlprefix}{URL }
\expandafter\ifx\csname urlstyle\endcsname\relax
  \providecommand{\doi}[1]{doi:\discretionary{}{}{}#1}\else
  \providecommand{\doi}{doi:\discretionary{}{}{}\begingroup
  \urlstyle{rm}\Url}\fi

\bibitem[{Abdollahi and Pradhan(2021)}]{Abdollahi2021}
Abdollahi, A. and Pradhan, B. (2021).
\newblock {Urban Vegetation Mapping from Aerial Imagery Using Explainable AI
  (XAI)}.
\newblock \emph{Sensors}, 21(14), 4738.
\newblock \doi{10.3390/s21144738}.

\bibitem[{Ameli et~al.(2022)Ameli, Aremanda, Friess, and Landis}]{Ameli2022}
Ameli, Z., Aremanda, Y., Friess, W.A., and Landis, E.N. (2022).
\newblock {Impact of UAV Hardware Options on Bridge Inspection Mission
  Capabilities}.
\newblock \emph{Drones}, 6(3), 64.
\newblock \doi{10.3390/drones6030064}.

\bibitem[{Araar et~al.(2015)Araar, Aouf, and Dietz}]{Araar2015}
Araar, O., Aouf, N., and Dietz, J.L.V. (2015).
\newblock {Power pylon detection and monocular depth estimation from inspection
  UAVs}.
\newblock \emph{Industrial Robot}, 42(3), 200--213.
\newblock \doi{10.1108/IR-11-2014-0419}.

\bibitem[{Badrinarayanan et~al.(2017)Badrinarayanan, Kendall, and
  Cipolla}]{Badrinarayanan2017}
Badrinarayanan, V., Kendall, A., and Cipolla, R. (2017).
\newblock {SegNet: A Deep Convolutional Encoder-Decoder Architecture for Image
  Segmentation}.
\newblock \emph{IEEE Transactions on Pattern Analysis and Machine
  Intelligence}, 39(12), 2481--2495.
\newblock \doi{10.1109/TPAMI.2016.2644615}.

\bibitem[{Bay et~al.(2005)Bay, Ferraris, and Van~Gool}]{Bay2005}
Bay, H., Ferraris, V., and Van~Gool, L. (2005).
\newblock Wide-baseline stereo matching with line segments.
\newblock In \emph{2005 IEEE Computer Society Conference on Computer Vision and
  Pattern Recognition (CVPR'05)}, volume~1, 329--336.
\newblock \doi{10.1109/CVPR.2005.375}.

\bibitem[{Bolourian and Hammad(2020)}]{Bolourian2020}
Bolourian, N. and Hammad, A. (2020).
\newblock {LiDAR-equipped UAV path planning considering potential locations of
  defects for bridge inspection}.
\newblock \emph{Automation in Construction}, 117(May), 103250.
\newblock \doi{10.1016/j.autcon.2020.103250}.

\bibitem[{Bono et~al.(2022)Bono, D'alfonso, Fedele, Filice, and
  Natalizio}]{Bono2022}
Bono, A., D'alfonso, L., Fedele, G., Filice, A., and Natalizio, E. (2022).
\newblock {Path Planning and Control of a UAV Fleet in Bridge Management
  Systems}.
\newblock \emph{Remote Sensing}, 14(8), 1858.
\newblock \doi{10.3390/rs14081858}.

\bibitem[{Bryant and Cios(2018)}]{Bryant2018}
Bryant, A. and Cios, K. (2018).
\newblock {Rnn-dbscan: A density-based clustering algorithm using reverse
  nearest neighbor density estimates}.
\newblock \emph{IEEE Transactions on Knowledge and Data Engineering}, 30(6),
  1109--1121.
\newblock \doi{10.1109/TKDE.2017.2787640}.

\bibitem[{Burt et~al.(2019)Burt, Disney, and Calders}]{Burt2019}
Burt, A., Disney, M., and Calders, K. (2019).
\newblock {Extracting individual trees from lidar point clouds using treeseg}.
\newblock \emph{Methods in Ecology and Evolution}, 10(3), 438--445.
\newblock \doi{10.1111/2041-210X.13121}.

\bibitem[{Car et~al.(2020)Car, Markovic, Ivanovic, Orsag, and Bogdan}]{Car2020}
Car, M., Markovic, L., Ivanovic, A., Orsag, M., and Bogdan, S. (2020).
\newblock {Autonomous Wind-Turbine Blade Inspection Using LiDAR-Equipped
  Unmanned Aerial Vehicle}.
\newblock \emph{IEEE Access}, 8, 131380--131387.
\newblock \doi{10.1109/ACCESS.2020.3009738}.

\bibitem[{Cheng et~al.(2014)Cheng, Tong, Wang, and Li}]{Cheng2014}
Cheng, L., Tong, L., Wang, Y., and Li, M. (2014).
\newblock {Extraction of Urban Power Lines from Vehicle-Borne LiDAR Data}.
\newblock \emph{Remote Sensing}, 6(4), 3302--3320.
\newblock \doi{10.3390/rs6043302}.

\bibitem[{Chi et~al.(2019)Chi, Lei, Shan, Wei, and Hao}]{Chi2019}
Chi, P., Lei, Y., Shan, S.S., Wei, Z., and Hao, D. (2019).
\newblock {Research on Power Line Segmentation and Tree Barrier Analysis}.
\newblock In \emph{2019 3rd International Conference on Electronic Information
  Technology and Computer Engineering (EITCE)}, 1395--1399. IEEE.
\newblock \doi{10.1109/EITCE47263.2019.9094966}.

\bibitem[{Corke and Khatib(2011)}]{corke2011robotics}
Corke, P. and Khatib, O. (2011).
\newblock \emph{Robotics, vision and control: fundamental algorithms in
  MATLAB}, volume~73.
\newblock Springer.

\bibitem[{Ester et~al.(1996)Ester, Kriegel, Sander, and Xu}]{Easter1996}
Ester, M., Kriegel, H.P., Sander, J., and Xu, X. (1996).
\newblock A density-based algorithm for discovering clusters in large spatial
  databases with noise.
\newblock In \emph{Proceedings of the Second International Conference on
  Knowledge Discovery and Data Mining}, KDD'96, 226–231. AAAI Press.

\bibitem[{Gao et~al.(2019)Gao, Yang, Li, Shen, Wang, Tian, Wang, and
  Liang}]{Gao2019}
Gao, Z., Yang, G., Li, E., Shen, T., Wang, Z., Tian, Y., Wang, H., and Liang,
  Z. (2019).
\newblock {Insulator Segmentation for Power Line Inspection Based on Modified
  Conditional Generative Adversarial Network}.
\newblock \emph{Journal of Sensors}, 2019, 1--8.
\newblock \doi{10.1155/2019/4245329}.

\bibitem[{Guan et~al.(2016)Guan, Yu, Li, Ji, and Zhang}]{Guan2016}
Guan, H., Yu, Y., Li, J., Ji, Z., and Zhang, Q. (2016).
\newblock {Extraction of power-transmission lines from vehicle-borne lidar
  data}.
\newblock \emph{International Journal of Remote Sensing}, 37(1), 229--247.
\newblock \doi{10.1080/01431161.2015.1125549}.

\bibitem[{Guo et~al.(2015)Guo, Huang, Zhang, and Sohn}]{Guo2015}
Guo, B., Huang, X., Zhang, F., and Sohn, G. (2015).
\newblock {Classification of airborne laser scanning data using JointBoost}.
\newblock \emph{ISPRS Journal of Photogrammetry and Remote Sensing}, 100,
  71--83.
\newblock \doi{10.1016/j.isprsjprs.2014.04.015}.

\bibitem[{Guo et~al.(2016)Guo, Li, Huang, and Wang}]{Guo2016}
Guo, B., Li, Q., Huang, X., and Wang, C. (2016).
\newblock {An improved method for power-line reconstruction from point cloud
  data}.
\newblock \emph{Remote Sensing}, 8(1), 36.

\bibitem[{Huang et~al.(2021)Huang, Du, and Shi}]{Huang2021}
Huang, Y., Du, Y., and Shi, W. (2021).
\newblock {Fast and accurate power line corridor survey using spatial line
  clustering of point cloud}.
\newblock \emph{Remote Sensing}, 13(8).
\newblock \doi{10.3390/rs13081571}.

\bibitem[{Isola et~al.(2017)Isola, Zhu, Zhou, and Efros}]{Isola2017}
Isola, P., Zhu, J.Y., Zhou, T., and Efros, A.A. (2017).
\newblock {Image-to-image translation with conditional adversarial networks}.
\newblock In \emph{Proceedings - 30th IEEE Conference on Computer Vision and
  Pattern Recognition, CVPR 2017}, volume 2017-January, 5967--5976. IEEE.
\newblock \doi{10.1109/CVPR.2017.632}.

\bibitem[{Jordan et~al.(2018)Jordan, Moore, Hovet, Box, Perry, Kirsche, Lewis,
  and Tse}]{Jordan2018}
Jordan, S., Moore, J., Hovet, S., Box, J., Perry, J., Kirsche, K., Lewis, D.,
  and Tse, Z.T.H. (2018).
\newblock {State-of-the-art technologies for UAV inspections}.
\newblock \emph{IET Radar, Sonar and Navigation}, 12(2), 151--164.
\newblock \doi{10.1049/iet-rsn.2017.0251}.

\bibitem[{Jwa et~al.(2009)Jwa, Sohn, and Kim}]{Jwa2009}
Jwa, Y., Sohn, G., and Kim, H. (2009).
\newblock Automatic 3d powerline reconstruction using airborne lidar data.
\newblock \emph{IAPRS}, 38, 105--110.

\bibitem[{Li et~al.(2022)Li, Luo, Xiao, Chen, Wang, and Li}]{Li2022}
Li, W., Luo, Z., Xiao, Z., Chen, Y., Wang, C., and Li, J. (2022).
\newblock {A GCN-Based Method for Extracting Power Lines and Pylons from
  Airborne LiDAR Data}.
\newblock \emph{IEEE Transactions on Geoscience and Remote Sensing}, 60.
\newblock \doi{10.1109/TGRS.2021.3076107}.

\bibitem[{Liang et~al.(2011)Liang, Zhang, Deng, Liu, and Shi}]{Liang2011}
Liang, J., Zhang, J., Deng, K., Liu, Z., and Shi, Q. (2011).
\newblock {A New Power-Line Extraction Method Based on Airborne LiDAR Point
  Cloud Data}.
\newblock In \emph{2011 International Symposium on Image and Data Fusion},
  1--4. IEEE.
\newblock \doi{10.1109/ISIDF.2011.6024293}.

\bibitem[{Lin(2016)}]{Lin2016}
Lin, X. (2016).
\newblock A method for powerline lidar point cloud segmentation using k-means
  clustering of a feature space.
\newblock \emph{Science of Surveying and Mapping}, 41, 60--63.

\bibitem[{Liu et~al.(2019)Liu, Wang, and Liu}]{Liu2019}
Liu, Z., Wang, X., and Liu, Y. (2019).
\newblock {Application of Unmanned Aerial Vehicle Hangar in Transmission Tower
  Inspection Considering the Risk Probabilities of Steel Towers}.
\newblock \emph{IEEE Access}, 7, 159048--159057.
\newblock \doi{10.1109/ACCESS.2019.2950682}.

\bibitem[{Lodha et~al.(2007)Lodha, Fitzpatrick, and Helmbold}]{Lodha2007}
Lodha, S.K., Fitzpatrick, D.M., and Helmbold, D.P. (2007).
\newblock {Aerial lidar data classification using expectation-maximization}.
\newblock In \emph{Vision Geometry XV}, volume 6499, 64990L. SPIE.
\newblock \doi{10.1117/12.714713}.

\bibitem[{McInnes and Healy(2017)}]{McInnes2017}
McInnes, L. and Healy, J. (2017).
\newblock {Accelerated Hierarchical Density Based Clustering}.
\newblock In \emph{2017 IEEE International Conference on Data Mining Workshops
  (ICDMW)}, volume 2017-Novem, 33--42. IEEE.
\newblock \doi{10.1109/ICDMW.2017.12}.

\bibitem[{McLaughlin(2006)}]{McLaughlin2006}
McLaughlin, R. (2006).
\newblock Extracting transmission lines from airborne lidar data.
\newblock \emph{IEEE Geoscience and Remote Sensing Letters}, 3(2), 222--226.
\newblock \doi{10.1109/LGRS.2005.863390}.

\bibitem[{Munir et~al.(2019)Munir, Awrangjeb, Stantic, Lu, and
  Islam}]{Munir2019}
Munir, N., Awrangjeb, M., Stantic, B., Lu, G., and Islam, S. (2019).
\newblock {Voxel-based extraction of individual pylons and wires from LiDAR
  point cloud data}.
\newblock \emph{ISPRS Annals of the Photogrammetry, Remote Sensing and Spatial
  Information Sciences}, 4(4/W8), 91--98.
\newblock \doi{10.5194/isprs-annals-IV-4-W8-91-2019}.

\bibitem[{Nurunnabi et~al.(2021)Nurunnabi, Teferle, Li, Lindenbergh, and
  Parvaz}]{Nurunnabi2021}
Nurunnabi, A., Teferle, F.N., Li, J., Lindenbergh, R.C., and Parvaz, S. (2021).
\newblock {Investigation of PointNet for semantic segmentation of large-scale
  outdoor point clouds}.
\newblock \emph{The International Archives of the Photogrammetry, Remote
  Sensing and Spatial Information Sciences}, XLVI-4/W5-(4/W5-2021), 397--404.

\bibitem[{Qiao et~al.(2020)Qiao, Sun, and Zhang}]{Qiao2020}
Qiao, S., Sun, Y., and Zhang, H. (2020).
\newblock {Deep learning based electric pylon detection in remote sensing
  images}.
\newblock \emph{Remote Sensing}, 12(11), 1--29.
\newblock \doi{10.3390/rs12111857}.

\bibitem[{Ritter and Benger(2012)}]{Ritter2012}
Ritter, M. and Benger, W. (2012).
\newblock {Reconstructing power cables from LIDAR data using Eigenvector
  streamlines of the point distribution tensor field}.
\newblock \emph{Journal of WSCG}, 20(3), 223--230.

\bibitem[{Shang and Shen(2018)}]{Shang2018}
Shang, Z. and Shen, Z. (2018).
\newblock {Real-time 3D reconstruction on construction site using visual SLAM
  and UAV}.
\newblock \emph{Construction Research Congress 2018: Construction Information
  Technology - Selected Papers from the Construction Research Congress 2018},
  2018-April, 305--315.
\newblock \doi{10.1061/9780784481264.030}.

\bibitem[{Shi et~al.(2018)Shi, Kang, Lin, Liu, and Chen}]{Shi2018}
Shi, Z., Kang, Z., Lin, Y., Liu, Y., and Chen, W. (2018).
\newblock {Automatic recognition of pole-like objects from mobile laser
  scanning point clouds}.
\newblock \emph{Remote Sensing}, 10(12), 1--23.
\newblock \doi{10.3390/rs10121891}.

\bibitem[{Sohn et~al.(2012)Sohn, Jwa, and Kim}]{Sohn2012}
Sohn, G., Jwa, Y., and Kim, H.B. (2012).
\newblock {Automatic power-line scence classification and reconstruction using
  airborn LiDAR data}.
\newblock \emph{ISPRS Annals of the Photogrammetry, Remote Sensing and Spatial
  Information Sciences}, I-3(September), 167--172.
\newblock \doi{10.5194/isprsannals-I-3-167-2012}.

\bibitem[{Teng et~al.(2017)Teng, Zhou, Li, Wu, Li, Meng, Zhou, and
  Ma}]{Teng2017}
Teng, G.E., Zhou, M., Li, C.R., Wu, H.H., Li, W., Meng, F.R., Zhou, C.C., and
  Ma, L. (2017).
\newblock {Mini-UAV LIDAR for power line inspection}.
\newblock \emph{International Archives of the Photogrammetry, Remote Sensing
  and Spatial Information Sciences - ISPRS Archives}, 42(2W7), 297--300.
\newblock \doi{10.5194/isprs-archives-XLII-2-W7-297-2017}.

\bibitem[{Wang et~al.(2019)Wang, Chen, Hua, and Zheng}]{Wang2019}
Wang, L., Chen, Z., Hua, D., and Zheng, Z. (2019).
\newblock {Semantic Segmentation of Transmission Lines and Their Accessories
  Based on UAV-Taken Images}.
\newblock \emph{IEEE Access}, 7, 80829--80839.
\newblock \doi{10.1109/ACCESS.2019.2923024}.

\bibitem[{Wang et~al.(2018)Wang, Peethambaran, and Chen}]{Wang2018}
Wang, R., Peethambaran, J., and Chen, D. (2018).
\newblock {LiDAR Point Clouds to 3-D Urban Models : A Review}.
\newblock \emph{IEEE Journal of Selected Topics in Applied Earth Observations
  and Remote Sensing}, 11(2), 606--627.
\newblock \doi{10.1109/JSTARS.2017.2781132}.

\bibitem[{Wang et~al.(2017)Wang, Chen, Liu, Zheng, Li, and Li}]{Wang2017}
Wang, Y., Chen, Q., Liu, L., Zheng, D., Li, C., and Li, K. (2017).
\newblock {Supervised Classification of Power Lines from Airborne LiDAR Data in
  Urban Areas}.
\newblock \emph{Remote Sensing}, 9(8), 771.
\newblock \doi{10.3390/rs9080771}.

\bibitem[{Wang et~al.(2021)Wang, Yang, Sheng, and Shen}]{Wang2021}
Wang, Z., Yang, L., Sheng, Y., and Shen, M. (2021).
\newblock {Pole-like objects segmentation and multiscale classification-based
  fusion from mobile point clouds in road scenes}.
\newblock \emph{Remote Sensing}, 13(21).
\newblock \doi{10.3390/rs13214382}.

\bibitem[{Yang et~al.(2020)Yang, Fan, Liu, Li, Peng, and Liang}]{Yang2020}
Yang, L., Fan, J., Liu, Y., Li, E., Peng, J., and Liang, Z. (2020).
\newblock {A Review on State-of-the-Art Power Line Inspection Techniques}.
\newblock \emph{IEEE Transactions on Instrumentation and Measurement}, 69(12),
  9350--9365.
\newblock \doi{10.1109/TIM.2020.3031194}.

\bibitem[{Yermo et~al.(2019)Yermo, Mart{\'{i}}nez, Lorenzo, Vilari{\~{n}}o,
  Cabaleiro, Pena, and Rivera}]{Yermo2019}
Yermo, M., Mart{\'{i}}nez, J., Lorenzo, O.G., Vilari{\~{n}}o, D.L., Cabaleiro,
  J.C., Pena, T.F., and Rivera, F.F. (2019).
\newblock {Automatic detection and characterisation of power lines and their
  surroundings using lidar data}.
\newblock \emph{International Archives of the Photogrammetry, Remote Sensing
  and Spatial Information Sciences - ISPRS Archives}, 42(2/W13), 1161--1168.
\newblock \doi{10.5194/isprs-archives-XLII-2-W13-1161-2019}.

\bibitem[{Zhao et~al.(2016)Zhao, Xu, Qi, Liu, and Zhang}]{Zhao2016}
Zhao, Z., Xu, G., Qi, Y., Liu, N., and Zhang, T. (2016).
\newblock {Multi-patch deep features for power line insulator status
  classification from aerial images}.
\newblock In \emph{2016 International Joint Conference on Neural Networks
  (IJCNN)}, volume 2016-Octob, 3187--3194. IEEE.
\newblock \doi{10.1109/IJCNN.2016.7727606}.

\end{thebibliography}
\end{document}